\documentclass[letterpaper]{article}
\usepackage{uai2018}
\usepackage[margin=1in]{geometry}

\PassOptionsToPackage{numbers, compress}{natbib}

\usepackage{times}
\usepackage{amsmath}
\usepackage{amssymb}
\usepackage{url}
\usepackage{graphicx}
\usepackage{natbib}
\usepackage{hyperref} 

\title{Latent Space Cartography: Generalised Metric-Inspired Measures and Measure-Based Transformations for Generative Models}

\author{
  Max F. Frenzel\\
  Cogent Labs\\
  Tokyo, Japan\\
  \And
  Bogdan Teleaga\thanks{Work performed during an internship at Cogent Labs.}\\
  Cogent Labs\\
  Tokyo, Japan\\
   \And
  Asahi Ushio\\
  Cogent Labs\\
  Tokyo, Japan\\
}

\begin{document}

\maketitle

\begin{abstract}
Deep generative models are universal tools for learning data distributions on high dimensional data spaces via a mapping to lower dimensional latent spaces. We provide a study of latent space geometries and extend and build upon previous results on Riemannian metrics. We show how a class of heuristic measures gives more flexibility in finding meaningful, problem-specific distances, and how it can be applied to diverse generator types such as autoregressive generators commonly used in e.g. language and other sequence modeling. We further demonstrate how a diffusion-inspired transformation previously studied in cartography can be used to smooth out latent spaces, stretching them according to a chosen measure. In addition to providing more meaningful distances directly in latent space, this also provides a unique tool for novel kinds of data visualizations. We believe that the proposed methods can be a valuable tool for studying the structure of latent spaces and learned data distributions of generative models.

\end{abstract}

%\begin{keywords}%
%  List of keywords%
%\end{keywords}

\section{Introduction\label{sec:Introduction}}

Deep generative models such as Variational Autoencoders (VAEs) \cite{Kingma2013-cu, Burda2015-qb} and Generative Adversarial Networks (GANs) \cite{NIPS2014_5423} have played a prominent role in the advancement of unsupervised learning. While the details of the architectures vary widely, the general objective is the same across generative models. Given a set of $N$ observations $\mathbf{X} = \{\mathbf{x}^{(1)}, ... , \mathbf{x}^{(N)}\}$ in observation space $\mathbf{\mathcal{X}}$ with dimension $D_x$, we want to learn a function that can model the data via the latent variables $\mathbf{Z} = \{\mathbf{z}^{(1)}, ... , \mathbf{z}^{(N)}\}$ in the latent space $\mathbf{\mathcal{Z}}$, which is usually of much lower dimension $D_z \ll D_x$. The (potentially stochastic) generator function $g : \mathcal{Z}\rightarrow \mathcal{X}$ allows us to map from an arbitrary latent variable $\mathbf{z}$ to its corresponding observation in data space $\mathbf{x}=g(\mathbf{z})$.\\
Many prominent applications of generative models such as clustering, comparisons of semantic similarity, and data interpolations rely heavily on the notion of distance in latent space. However, there is generally no guarantee that distances in latent space represent a meaningful measure. The notion of meaningful distances itself can be hard to define and highly problem-specific. The manifold hypothesis asserts that the observations in data space actually lie on a low dimensional manifold. This is the key property exploited by generative models. However, while observations in $\mathbf{\mathcal{X}}$ might be extremely sparse in certain regions, the training objectives of most generative models promote a densely packed latent space. This can lead to rather dissimilar observations being embedded in close proximity.\\
This problem as well as the importance of meaningful distances have led to a growing interest in the geometry of latent spaces. Several groups have independently introduced the idea of applying Riemannian geometry to define a metric on the latent space \cite{Chen2017-pk, DBLP:journals/corr/abs-1711-08014, Arvanitidis2017-oi, Hadjeres2017-ws, featuremetric, Peste}, which allows for concepts such as geodesics that give distances and shortest paths which more closely reflect the data. This requires a meaningful and differentiable metric in $\mathbf{\mathcal{X}}$. However, this limits the potential metrics that can be used and excludes entire classes of models such as language models whose generators rely on repeated sampling.\\
In this work we extend and build on previous results in several aspects. In section \ref{sec:Metrics}, after briefly reviewing the general idea of a metric on latent space as well as previous work on Riemannian metrics and their limitations, we introduce an easy to implement method for approximating a wide range of heuristic metrics and metric-inspired measures. These quantities, while not as rigorous as Riemannian metrics and in some cases lacking certain desirable properties, can be applied to any type of generator function and can be precisely engineered based on the specific problem to be solved, offering considerably more flexibility. Section \ref{sec:Transform} introduces a diffusion based transformation and investigates how this transformation can be used to smooth out a latent space according to a particular measure, integrating meaningful distance information directly in the latent space itself. In section \ref{sec:Exp} we present a qualitative analysis of the measures induced by different architectures as well as different data types, and discuss how these results can be used for improved visualizations and how desired properties can be incorporated into the visualization by the choice of measure. We further show that given a good measure, one can find a transformed latent space which has more meaningful distances, and allows for improved interpolations and semantic clustering. Finally we conclude in section \ref{sec:Conc}.

\section{Latent Space Metrics\label{sec:Metrics}}
A metric defines a notion of distance between any pair of points in a space. This distance is non-negative, symmetric, satisfies the triangle-inequality, and vanishes iff the two points coincide. The most frequently used metric in many applications is the Euclidean or $L_2$ metric which is given by the length of the straight line connecting two points. The use of this metric comes with the implicit assumption that the underlying space is Euclidean and has no distortions or curvature. However, despite the widespread use of this metric in applications that rely on distances in latent space, there is in general no guarantee that latent spaces are actually flat Euclidean spaces.\\
On the contrary, the training objectives of most generative models naturally encourage the space to be stretched in some areas and compressed in others. The evidence lower bound in VAEs for example contains a term that encourages the learned approximate posterior distribution $q(\mathbf{z} | \mathbf{x})$ to match a (usually Gaussian) prior $p(\mathbf{z})$ via minimization of the KL-divergence $D_{KL} (q(\mathbf{z} | \mathbf{x}) || p(\mathbf{z}))$, whose asymmetric nature encourages the posterior to completely fill the prior, $q(\mathbf{z} | \mathbf{x})$ not having any low density regions where $p(\mathbf{z})$ has high density. Qualitatively, we encourage our models to learn a data distribution that is free of holes. However, unless we also have data that is uniformly distributed, this naturally leads to distortions in latent space. A small volume occupied by e.g. a category boundary in latent space may actually correspond to a vast empty volume in data space. Trying to match the data to a Gaussian prior also induces a higher density near the origin. We can expect that a segment of distance $\Delta z$ that is close to the origin covers a larger variety of data than a segment of the same length that lies towards the edge of the distribution (c.f. Fig {\ref{fig:main}). Thus the Euclidean metric is in general inadequate to represent meaningful distances between latent variables.

\subsection{Riemannian Metrics and their Limitations}
This problem has inspired a number of recent investigations aiming to use ideas from Riemannian geometry to define a more suitable metric. In particular, the authors in \cite{Chen2017-pk, DBLP:journals/corr/abs-1711-08014, Arvanitidis2017-oi} advocate the use of a Riemannian metric instead of $L_2$, treating the latent space as a Riemannian manifold. This allows one to replace distances in latent space, where a readily available notion of distance is generally lacking, with distances in observation space for which it is assumed that we do have a meaningful measure of distance.\\
For a detailed discussion of Riemannian manifolds and metrics we refer the reader to the original literature \cite{Chen2017-pk, DBLP:journals/corr/abs-1711-08014, Arvanitidis2017-oi}, but the general idea is to consider the Jacobian $\mathbf{J}$ with respect to some map $f : \mathcal{Z}\rightarrow \mathcal{H}\subseteq\mathbb{R}^{D_h}$. From the Jacobian we can get the metric tensor $\mathbf{M}=\mathbf{J}(\mathbf{z})^T\mathbf{J}(\mathbf{z})$, a symmetric positive definite matrix that encodes local curvature of the space. Related to this is the associated Riemannian measure\footnote{The Riemannian measure is also sometimes referred to as the magnification factor or volume element. In the following we will be using the terms measure, magnification factor, and volume element interchangeably.} $m_{RM}(\mathbf{z}) = \sqrt{\det \mathbf{J}(\mathbf{z})^T\mathbf{J}(\mathbf{z})}$, which quantifies how much volume in $\mathcal{H}$ an infinitesimal volume around $\mathbf{z}$ occupies. It essentially defines a density distribution over $\mathcal{Z}$. The non-uniformity of this distribution gives us a notion of how distorted $\mathcal{Z}$ is. Assuming isotropy, we can consider $m_{RM}(\mathbf{z})$ as a multiplicative factor applied to an infinitesimal line segment $\mathbf{dz}$ passing through $\mathbf{z}$. The more volume in $\mathcal{H}$ an infinitesimal unit cell around $\mathbf{z}$ corresponds to, and thus the larger $m_{RM}(\mathbf{z})$, the more distance we should assign to the segment $\mathbf{dz}$. We will also use this idea as the basis for our heuristic measures introduced below which are not necessarily derived from a Jacobian. While the assumption of isotropy is very strong and usually does not hold, and the stretching of the line segment $\mathbf{dz}$ should thus also depend on its direction, we find that this assumption still leads to useful results as we shall show below.\\
The original papers introducing the idea defined the Jacobian with respect to the generator function $g$ mapping to observation space, i.e. $f=g$  and $\mathcal{H}=\mathcal{X}$ such that $J_{i,j} = \frac{\partial x_i}{\partial z_j}$. However, as pointed out in \cite{featuremetric} this only provides a meaningful metric if Euclidean distances in $\mathbf{\mathcal{X}}$ are meaningful. For example for images it is highly questionable whether the $L_2$ metric provides a meaningful measure of semantic distance. The authors suggest an alternative metric which is not defined on the final output space, but on some intermediate activation layer in the generator. Since hidden units in intermediate layers tend to represent certain features, defining the Jacobian with respect to their activations should provide distances that capture semantic ideas rather than linear interpolations of the data.\\
In any applications of the above ideas, it is crucial to have a meaningful and tractable metric on $\mathcal{H}$ informed by the data which we want to pull back into the latent space. In some cases such as the simulation of a pendulum in \cite{Chen2017-pk} we do indeed have a very meaningful metric on the data, namely the angle of the pendulum. But in other cases, such as for images, we might need to find a less obvious metric. In general, what metric  or measure we find ``meaningful'' might strongly depend on the specific problem we are trying to solve. Hence we would like to be as flexible as possible in the choice of metrics we can use. However, calculating the Riemannian metric requires the generator function $g$ (or more generally the map $f$) to be differentiable and smooth. This excludes generators which use sampling procedures in the generation process, such as the decoders of most \textit{seq2seq} models \cite{DBLP:journals/corr/ChoMGBSB14, DBLP:journals/corr/BahdanauCB14}. 

\subsection{Universal Heuristic Measures}
To circumvent these limitations we propose an approximate sampling based method that allows to both easily approximate the Riemannian metric to arbitrary precision, as well as provides large freedom in the design of other heuristic metrics and magnification factors for particular problems. This method can be applied to arbitrary generator functions, as well as more complex functions defined on the output of the generator or the latent space itself. Specifically we can consider an arbitrary function $f : \mathcal{Z}\rightarrow \mathcal{H}$, which maps a point $\mathbf{z}$ in $\mathcal{Z}$ to $\mathbf{h}:=f(\mathbf{z})$ in what we shall call the ``meaning space'' $\mathcal{H}\subseteq \mathbb{R}^{D_h}$, where a distance we deem meaningful is defined. We are free to choose an appropriate $\mathcal{H}$ and the corresponding $f$ based on the particular task we are trying to achieve.\\
To find the metric, or directly a heuristic measure, we begin by placing a square grid on the latent space that covers the entire embedded data $\mathbf{Z}$, with $N_i$ cells in the $i$th dimension. We then proceed by assigning each grid cell $\hat{c}$ in $\mathcal{Z}$ a characteristic $\mathbf{h}_{\hat{c}}$ in $\mathcal{H}$. Depending on the nature of $f$ and the grid resolution this can either simply be done by mapping each cell center $\mathbf{z}_{\hat{c}}$ to the corresponding point $\mathbf{h}_{\hat{c}} = f(\mathbf{z}_{\hat{c}})$, or, for stochastic generators, by sampling multiple $\mathbf{z}$ in $\hat{c}$ and defining $\mathbf{h}_{\hat{c}}$ as the average of the respective mappings. The resulting $(N_1\times ... \times N_{D_z} \times D_h)$-dimensional tensor $\mathbf{H}$ now forms the basis for calculating the metric or heuristic volume elements.

\subsubsection{Approximate Riemannian Metric\label{subsub:RM}}
Given the tensor $\mathbf{H}$ it is straightforward to use a simple finite-difference approximation for the Jacobian with respect to $f$, for example ${\mathbf{J}(\mathbf{z}_{i,j}) \approx 
\begin{bmatrix} \frac{\mathbf{h}_{i+1,j}-\mathbf{h}_{i-1,j}}{2\Delta z_1} & \frac{\mathbf{h}_{i,j+1}-\mathbf{h}_{i,j-1}}{2\Delta z_2} \end{bmatrix}}$ for $D_z=2$, 
where $\Delta z_1$ and $\Delta z_2$ are the cell widths in the two dimensions respectively. From this we can directly calculate an approximation to the measure ${m_{RM}(\mathbf{z}_{\hat{c}})=\sqrt{\det \mathbf{J}(\mathbf{z}_{\hat{c}})^T \mathbf{J}(\mathbf{z}_{\hat{c}})}}$.\\
If we choose $f$ to be the full generator function such that $\mathbf{h}=\mathbf{x}$, this provides an approximation to the metric considered in \cite{Chen2017-pk, DBLP:journals/corr/abs-1711-08014, Arvanitidis2017-oi}, whereas using a mapping to one of the intermediate layers in the generator recovers the feature based metric suggested in \cite{featuremetric}. 

\subsubsection{General Heuristic Measures and the Jensen-Shannon Measure for Autoregressive Generators\label{subsub:MFMetric}}
In addition to calculating $m_{RM}$ via a Jacobian, we can consider an arbitrary dissimilarity function $d_h(\cdot,\cdot)$ between points in $\mathcal{H}$ and directly define its corresponding heuristic measure $m_{d_h}$ on the grid as the average local dissimilarity under $d_h$ to its nearest neighbours. This approach, while sacrificing some directionality information, is extremely flexible and the potential applications are abundant. To illustrate the general idea and show one explicit application in the context of a common and highly relevant use case, let us consider the following specific example.\\
Autoregressive models are commonly used for sequence modeling tasks such as text or audio \cite{NIPS2014_5346, 476cadd89dec4e0ab01d9dc59e1222c7, Yang2017-dj, DBLP:journals/corr/OordDZSVGKSK16}. Their generators rely on repeated sampling from a distribution which gets conditioned on the past sampling history. In many such cases we do not even have a clear reasonably smooth metric on the observation space. For example for text, while there are measures such as edit distance \cite{jurafsky2000speech}, they are neither smooth nor do they capture semantic distance which is usually what we would like to capture.\\
We instead introduce an alternative measure which is not defined on the generated data itself, but the intermediate conditional distributions involved in generating it. Specifically, let us consider a generator representing a conditional language model with finite vocabulary \cite{476cadd89dec4e0ab01d9dc59e1222c7,Yang2017-dj}. At step $t$ in the generation process the generator produces a distribution over words $\mathbf{p}(w_t | \mathbf{z}, w_{t-1},...,{w_1})$, conditioned on both the latent variable $\mathbf{z}$ as well as the previous words. Assuming a total sentence length $T$, we now define our meaning function $f$ as the average over the intermediate word distributions
\begin{equation}\label{eq:hText}
\mathbf{h} = f(\mathbf{z}) := \frac{1}{T}\sum_{t=1}^T \mathbf{p}(w_t | \mathbf{z}, w_{t-1},...,{w_1}).
\end{equation}
This vector essentially captures the average word distribution associated with a point in latent space. Note that this is certainly not a perfect solution since word frequency alone is not enough to capture the full meaning of text, and it can assign the same $\mathbf{h}$ to different generator outputs. Despite these concerns, we did find this to be a useful quantity in practice. The question of better meaning capturing functions $f$ is an interesting direction for further research. One possible (albeit less easily interpretable) alternative could for example be the average hidden state of an LSTM decoder.\\
To arrive at a useful measure, we also need to define a suitable dissimilarity function $d_h(\cdot,\cdot)$. A natural choice in the cases where $\mathbf{h}$ represents probability distributions is the Jensen-Shannon distance ${JSD(h_1 || h_2) = \sqrt{ \frac{1}{2}D_{KL}(h_1 || \bar{h}) + \frac{1}{2}D_{KL}(h_2 || \bar{h}) }}$ where ${\bar{h} := \frac{1}{2}(h_1 + h_2)}$ \cite{MacKay:2002:ITI:971143}. With this choice ${d_h(\cdot,\cdot)=JSD(\cdot || \cdot)}$ we arrive at the Jensen-Shannon measure
\begin{equation}\label{eq:JSM}
m_{JSD}(\mathbf{z}_{\hat{c}}) := \frac{1}{n_{\hat{c}}}\sum_{<\hat{c}, \hat{c}'>} JSD(\mathbf{h}_{\hat{c}} || \mathbf{h}_{\hat{c}'}),
\end{equation}
where $<\hat{c}, \hat{c}'>$ denotes nearest neighbours and $n_{\hat{c}}$ is the number of cells bordering on $\hat{c}$. This measure quantifies how much the word distribution changes as we move through the unit cell around $\mathbf{z}_{\hat{c}}$.\\
As previously noted, this measure, just like the Riemannian measure derived from a metric tensor, has the drawback of lacking directional information, which might be crucial in certain applications, and an extension to the current approach that does not only provide a measure but a genuine metric could be a fruitful direction for further research. Despite this limitation, we still found these directionless quantities to be useful in practice. 

\subsubsection{Classifier Measures\label{subsub:CM}}
Another interesting type of metric can be found if categorical labels are available for the data $\mathbf{X}$. In this case we can train a classifier over the latent variables to predict the class $c$ given a latent variable $\mathbf{z}$, and use the resulting probability as the feature vector $\mathbf{h} = \mathbf{p}(c |\mathbf{z})$. Using again the Jensen-Shannon measure (\ref{eq:JSM}), $m_{JSD}(\mathbf{z})$ now captures how fast the class probabilities change in the vicinity of $\mathbf{z}$ and thus encodes information such as class boundaries. This can be interesting to get insights into the learned data distribution, as well as in conjunction with the transformation we introduce in section \ref{sec:Transform} to produce visualizations with clearly distinct clusters for each class.

\section{Latent Space Transformations\label{sec:Transform}}
\begin{figure*}[h]
  \centering
  \includegraphics[width=0.95\textwidth]{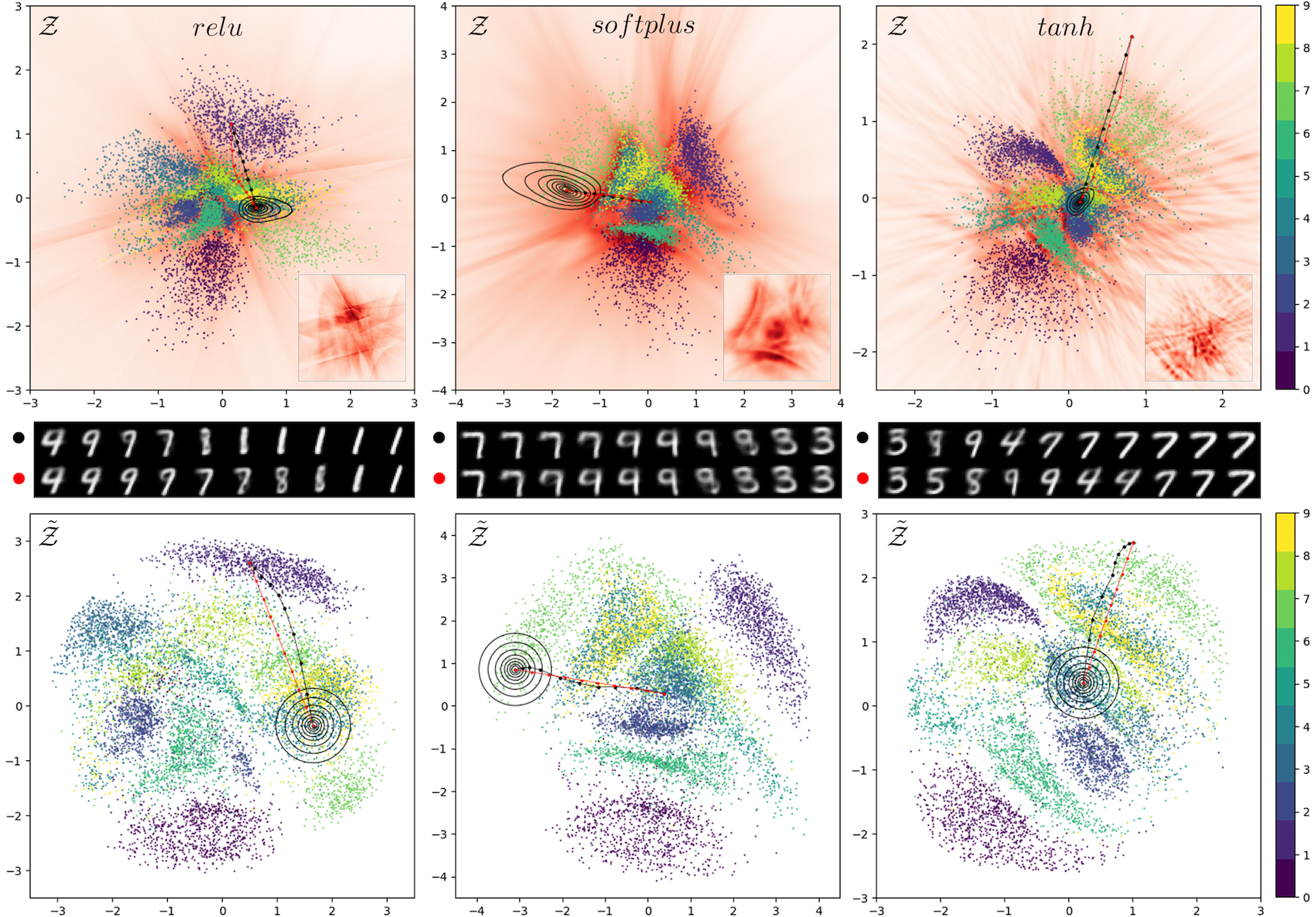}
  \caption{Learned MNIST data distributions and latent spaces with their respective measures, as well as data interpolations and the transformed latent spaces, for different activation functions. Top row: Original latent space $\mathcal{Z}$ with embedded validation data. The (for contrast) square rooted measure $\sqrt{m_{RM}}$ is shown in red. The inset shows $m_{RM}$ around the origin without embeddings. A straight path (black) and pseudo-geodesic (red) between two embeddings is shown, as well as equidistance lines around the starting point of the path. Middle row: Interpolations corresponding to the two paths (top: straight line; bottom: pseudo-geodesic). Bottom row: Embeddings, interpolation paths, and equidistance lines in the transformed space $\tilde{\mathcal{Z}}$.}
\label{fig:main}
\end{figure*}

The Riemannian metric and associated measure as well as the heuristic measures all quantify how much volume in an abstract meaning space $\mathcal{H}$ maps to each unit cell in latent space. As noted above, it essentially defines a density distribution over the latent space. This raises the question of whether we can find a transformation $T$ on the latent space that accounts for this unequal density and maps each point to a corresponding point ${\mathbf{z}\rightarrow \tilde{\mathbf{z}} := T(\mathbf{z}) \in \tilde{\mathcal{Z}}}$ , stretching the space in such a way as to equilibrate the density.\\
This problem has previously been studied in a seemingly unrelated domain: cartography. Specifically, cartograms \cite{b3ad6758e4074325b34caeadce5a91b5}, also known as density-equalizing maps, are maps in which the size of geographic regions is proportional to certain properties of that region, such as population or GDP. The most successful techniques for calculating the transformations underlying cartograms are inspired by physical diffusion processes \cite{Gastner2004-iv, Gastner2018-fr}, where we assume that the property of interest represents a particle density, and then allow the system to relax to its equilibrium state.\\
We can directly apply these methods to our present problem, essentially producing cartograms of latent spaces where the quantity of interest is the measure. The result is a bijective map $T$ from the original space $\mathcal{Z}$ to a stretched space $\tilde{\mathcal{Z}}$ in which unit volumes map to equal volumes in meaning space $\mathcal{H}$.\\
We refer the reader to the original cartogram literature \cite{Gastner2004-iv, Gastner2018-fr} for details on how to calculate the transformation $T$ given a density distribution\footnote{For all reported experiments we used the method proposed in \cite{Gastner2004-iv} and their open implementation which can be found at \url{http://www-personal.umich.edu/~mejn/cart/}. We also implemented \cite{Gastner2018-fr} but found the resulting maps to be less satisfying for metrics with very fine structure and high local gradients.}. In our scenario, treating the Riemannian or heuristic measure as a density distribution over the latent space, the resulting map $T$ is a discrete vector field over the grid defined on $\mathcal{Z}$, which maps each cell center $\mathbf{z}_{\hat{c}}$ to a corresponding point $\tilde{\mathbf{z}}_{\hat{c}}$. Using bilinear interpolation between cell centers we can find a continuous map for arbitrary points $\mathbf{z}\rightarrow \tilde{\mathbf{z}}$. It is also straightforward to (approximately) determine the inverse transformation $\tilde{T}$ such that $\tilde{T}(\tilde{\mathbf{z}}) = \mathbf{z}$.\\
Euclidean distances in $\tilde{\mathcal{Z}}$ give a much more faithful representation of semantic distance due to the direct incorporation of the meaure. However, note that due to the fact that we lose directionality information when considering the volume elements, straight lines in $\tilde{\mathcal{Z}}$ are in most cases not true geodesics, and we shall refer to them as \textit{pseudo-geodesics}. While not representing absolute shortest paths, they are still useful for distance comparisons, especially locally, as well as for data interpolations (c.f. Fig {\ref{fig:main}), and are trivial to compute. One could determine true geodesics for example by considering path integrals between points, but this comes at a significantly higher computational cost similar to approaches in \cite{Chen2017-pk} and \cite{Arvanitidis2017-oi} which rely on neural networks and solving a system of differential equations respectively to determine geodesics. We leave this to future investigations.\\
Similarly, finding a more advanced transformation that is not based on the measure, but the metric itself (assuming that a metric is available), and takes not just the local density but also directionality into account, is an interesting open question.

\section{Experiments\label{sec:Exp}}
Having introduced the methods for universal heuristic-based measures and the measure-smoothing transform, we now turn towards some explicit applications and give a qualitative study of how different model and data aspects affect latent spaces.
\subsection{MNIST}
We first consider the canonical example of MNIST images, which has also been studied in relation to latent space metrics by \cite{Chen2017-pk, DBLP:journals/corr/abs-1711-08014, Arvanitidis2017-oi}. For the VAE, we use a simple architecture consisting of two fully connected 512-unit layers each for encoder and decoder. We trained three separate models for the activation functions $relu(\cdot)$, $softplus(\cdot)$, and $tanh(\cdot)$ respectively and approximated their Riemannian measure $m_{RM}$ as described in \ref{subsub:RM}. Here and in all the following experiments we have used an $800\times800$-grid for the approximation. For comparison with previous work \cite{Chen2017-pk, DBLP:journals/corr/abs-1711-08014, Arvanitidis2017-oi} we also defined the Jacobian $\mathbf{J}$ with respect to the data space $\mathcal{X}$. Based on this metric we computed the transforms $T$ and applied it to the embeddings. We also performed data interpolations along straight paths in the original space, as well as the pseudo-geodesics. The results are shown in Fig. \ref{fig:main}.\\
Interestingly we find that the activation function used leaves a very strong imprint on the latent space. We confirmed via repeated experiments (not shown) that each activation function indeed leads to a very characteristic metric on the latent space. This shows that the proposed method can be a useful tool in the study of model architectures and activation functions. We also find that as expected, the interpolation along the pseudo-geodesic leads to smoother transitions. We note however that, as pointed out previously, smooth here only means linear interpolation between images due to the limited usefulness of the $L_2$-metric on the data, and not necessarily smooth in terms of semantic meaning.\\
Looking at the transformed space $\tilde{\mathcal{Z}}$, we note that despite the questionable adequacy of the $L_2$-metric on the data, the transformation achieves a visually nice separation of the data into more distinct clusters, as well as a very noticeable overall smoothing of the non-uniform Gaussian density induced by the VAE's prior. We believe that this is useful both in terms of more meaningful distances, as well as for improved visualizations. The increase in uniformity was further confirmed by calculating the entropy of the embeddings before and after the transformation, which showed a significant increase for all models considered.
\subsubsection{Improved Clustering}
\begin{figure}[h]
  \centering
  \includegraphics[width=0.45\textwidth]{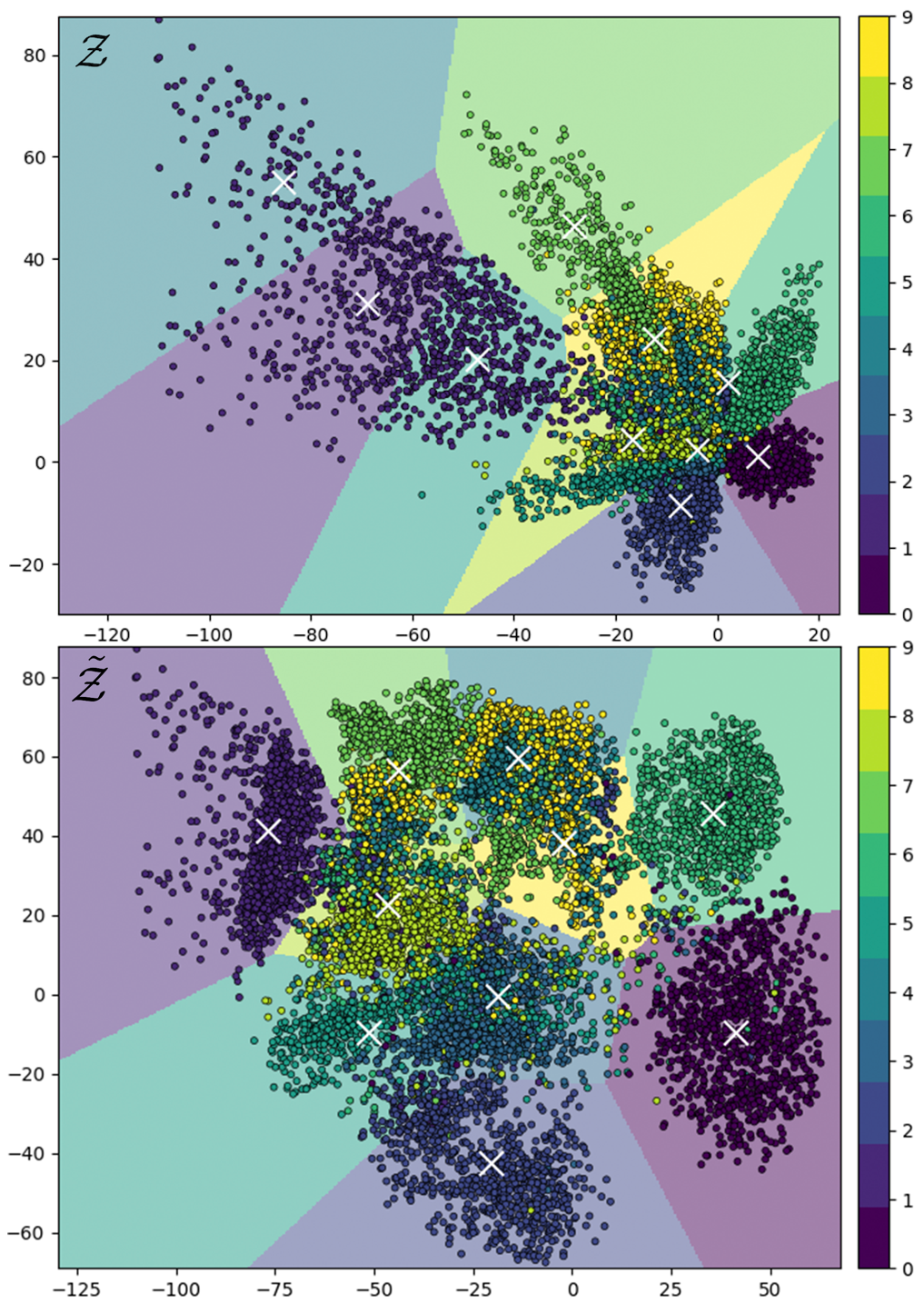}
  \caption{k-means clustering applied to the MNIST data distribution learned by a simple (non-variational) autoencoder in the original space $\mathcal{Z}$ (top) as well as the transformed space $\tilde{\mathcal{Z}}$ (bottom). The cluster-based classification F1-score improves from $0.414$ before to $0.569$ after the transformation.}
  \label{fig:clustering}
\end{figure}
To further study the cluster- and uniformity-improving properties of the transformation we trained a standard autoencoder (AE) with the same architecture as the above VAEs and with $relu$ activations. Due to the lack of regularization we expected the AE to have a highly distorted latent space. Fig. \ref{fig:clustering} shows that as expected, the transformation dramatically helps smooth out the distorted data distribution. We also performed k-means clustering based classification to get a more quantitative measure of the clustering, and found that the transformation lead to an improvement in F1-score from $0.414$ to $0.569$. We note however that repeating this same analysis for the VAEs only led to minor improvements. Given the questionable usefulness of the $L_2$-metric we did not necessarily expect it to lead to improved clustering at all. However, we believe that our method can be highly useful for clustering if one is able to find a measure which is more appropriate for capturing differences between the desired kinds of clusters (and again, our measures are flexible enough to be tweaked to accommodate certain desired properties and clusterings).
\subsubsection{Latent Space Distortions due to bad Training Data}
\begin{figure}[h]
  \centering
  \includegraphics[width=0.45\textwidth]{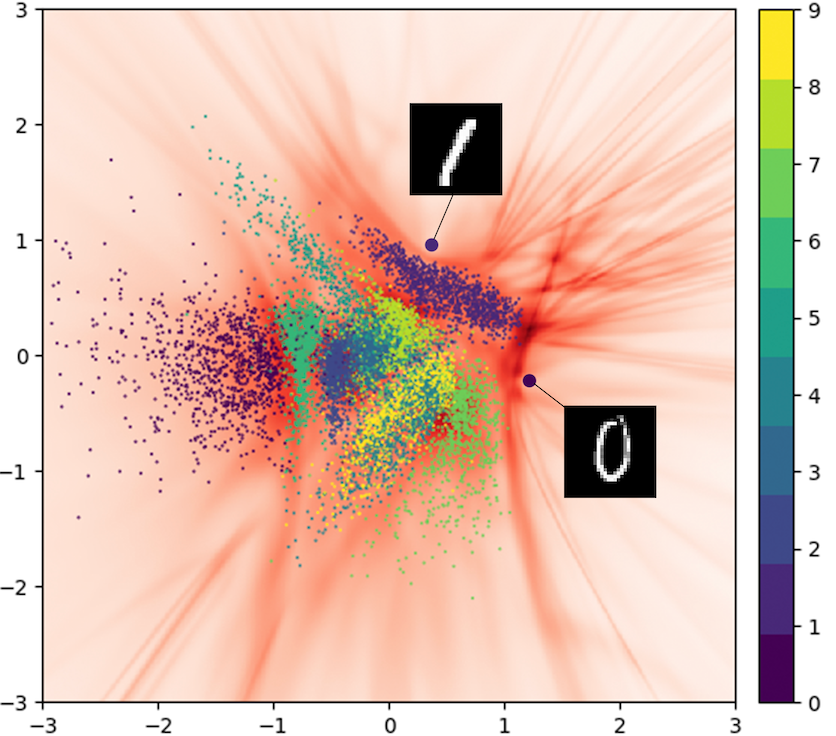}
  \caption{VAE trained on MNIST with two particular data points repeated $10,000$ times in the training data. The square rooted measure $\sqrt{m_{RM}}$ in red shows clearly how corrupted data (in this case repeated data) can lead to strong distortions and high curvature in latent space.}
\label{fig:repeated}
\end{figure}
Another interesting application is the study of the effects that corrupted training data have on the learned latent spaces. As a simple experiment we retrained the VAE with $softplus$ activation on a dataset in which we repeated two particular data points $10,000$ times. Fig. \ref{fig:repeated} very clearly shows the resulting distortion of the latent space, the VAE essentially reserving large and (particularly in the case of the ``0'') disconnected areas in latent space for the memorization of these samples. While being a simple toy example, this effectively demonstrates the importance of clean training data.
\subsubsection{Classifier Visualisations}
\begin{figure}[h]
  \centering
  \includegraphics[width=0.45\textwidth]{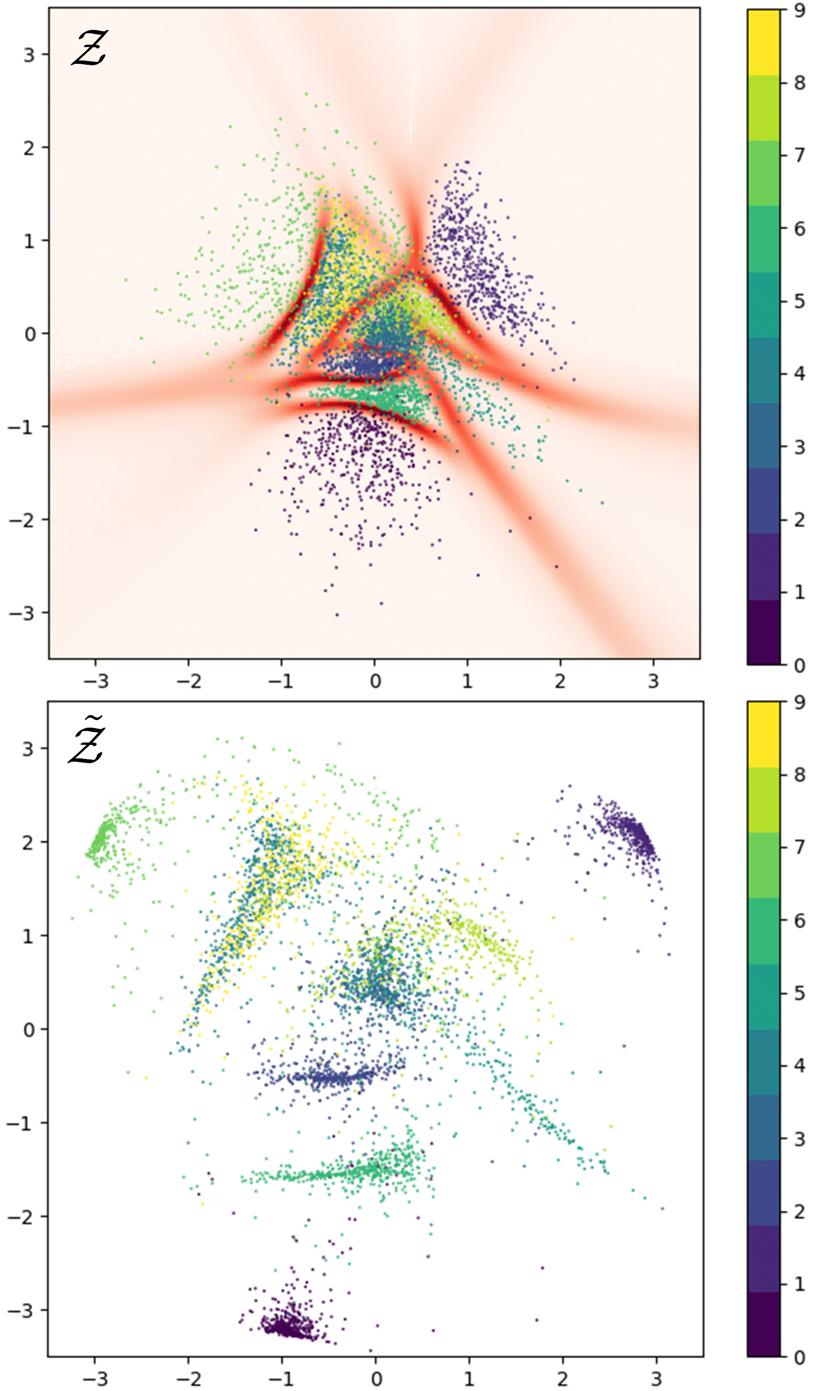}
  \caption{Top: Classifier based measure $m_{JSD}$ in red. Bottom: The same space after the transformation based on this measure .The strong class-separating effect that can be achieved through this measure and transformation is clearly visible.}
  \label{fig:classifier}
\end{figure}
To conclude our analysis of MNIST, we trained a simple MLP classifier \cite{scikit-learn} over the original embeddings of the $softplus$-VAE and calculated the classifier based measure $m_{JSD}$ as described in \ref{subsub:CM}. The measure and resulting transformation are shown in Fig. \ref{fig:classifier}. We can clearly see the class transitions learned by the classifier, as well as the resulting strong separation into clusters of unique classes. Again, while this is only a toy example, it shows how this type of measure can be used for producing visualizations that highlight distinct clusters of classes. One could imagine a more complicated dataset with hundreds of classes. If we want to highlight one or two particular classes, we could define a measure based only on those classes' respective classifier probabilities and use the resulting transformation to visually clearly separate out these classes from the remaining data. The strength of the desired effect can easily be controlled by applying a Gaussian blur to the metric.

\subsection{Natural Language\label{subsec:NLP}}
To show our methods applicability to a class of generators that was completely inaccessible to previous methods and to test our proposed Jensen-Shannon measure for autoregressive generators, we consider a language modeling \textit{seq2seq} VAE. In terms of architectures, we use exactly the same model and training procedure as proposed in \cite{Yang2017-dj}, with an LSTM encoder and dilated convolution based decoder.\\
We also introduce a new dataset that we found particularly interesting for language modeling (as well as classification) for practical applications. This dataset consists of consumer complaints about financial products collected by the US Government, along with various categorical labels such as product category\footnote{Data available at \url{https://www.consumerfinance.gov/data-research/consumer-complaints/}. This dataset is constantly updated, the version we used for experiments was retrieved on 23/05/2017.}. This dataset is interesting for its semantic richness, as well as labelled categories with fairly well defined semantic content.
\begin{figure*}[h]
  \centering
  \includegraphics[width=0.68\textwidth]{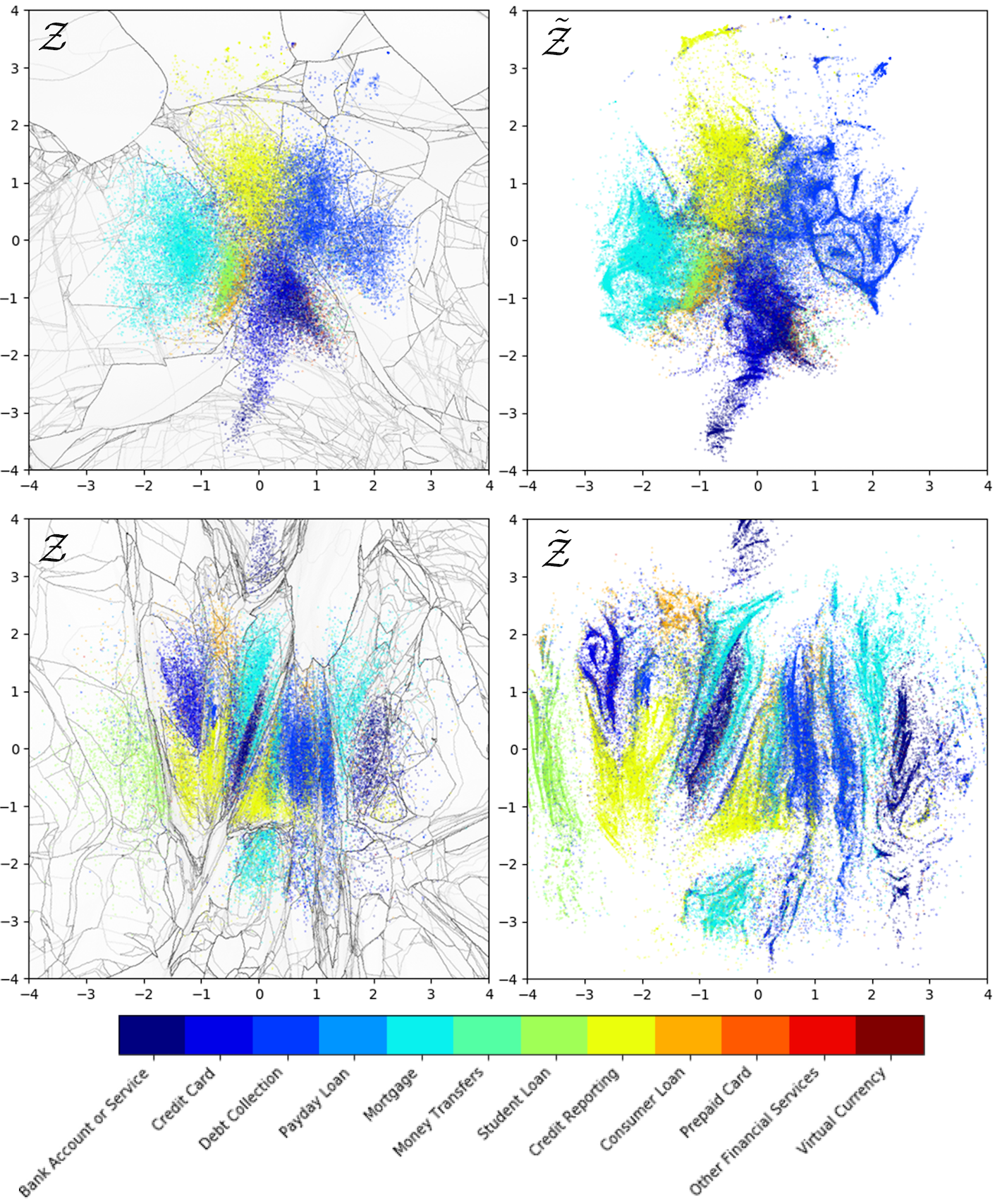}
  \caption{Average word distribution based Jensen-Shannon measure $m_{JSD}$ in black and embeddings (left), as well as transformed embeddings (right) for the Consumer Complaints dataset. Points are colored by product category. Top: \textit{seq2seq} VAE with $2$-dimensional latent space. Bottom: \textit{seq2seq} VAE with $D_z=128$ followed by compression to two dimensions (shown in the figure) via a second VAE.}
  \label{fig:text}
\end{figure*}\\
We first trained a model with $D_z=2$ and calculated the Jensen-Shannon measure and resulting transformation. Results are shown in Fig. \ref{fig:text}. Despite the small latent space dimension, the learned data distribution has surprisingly distinct regions for the different product categories. What is also remarkable is the complicated spider-web-like structure of the measure $m_{JSD}$. Regions of very smooth variation are enclosed by sharp boundaries that mark rapidly changing word distributions. This is reflected in the transformed space by strong clusters emerging in regions that were previously uniform. We can also see that the generator has learned a complicated function that extends well beyond the range of the actual data. To not give too much weight to the volume elements in these regions, we smoothly relaxed the measure to its average away from any data point before calculating the transformation.\\
To show that despite the methods' current computational cost which restricts them to low dimension they can still be useful for analysis of high dimensional latent spaces, we first trained a \textit{seq2seq} VAE with $D_z=128$ and then trained a second two-dimensional VAE on the learned embeddings, a simple compression VAE with two fully-connected layers for both encoder and decoder. We then calculated the JSD-measure on the two dimensional latent space by feeding the output of the compression VAE's generator as latent variable $\mathbf{z}$ to the generator of the \textit{seq2seq} VAE. Results are shown in the bottom half of Fig. \ref{fig:text}. We find that the higher dimensional VAE generally learns a much more distinct separation between the categories. Further, we see that the measure has the same web-like characteristics, although it could be argued that it looks qualitatively more complex. Remarkably, after applying the transformation, we find that the continuous distribution is split into semantically meaningful clusters, both between category boundaries, as well as within individual categories.\\
These results show both that the metric does indeed capture semantic meaning, as well as that this method could potentially be very useful for clustering beyond human-defined labels. A more detailed study of how the enclosed areas in the metric arise and how they can be used for defining clusters is another interesting area for future study. 

\section{Conclusion and Outlook\label{sec:Conc}}
We have introduced an easy to implement approximate method for the calculation of Riemannian metrics, as well as more general heuristic metrics and measures. Based on these measures, we introduced a density-equalizing transformation that allows to smooth out the latent space and rescale local distances according to certain desired properties. We used the proposed methods to analyze the effects of different types of architecture and data on the learned latent space.\\
The proposed methods in their present form certainly have drawbacks, particularly the exponential scaling with the latent space dimension $D_z$ which limits the applicability to low dimensions, as well as the lack of directionality information in the measure and resulting transformation. The ``un-distortion'' effect of the transformation is only perfect if the metric is isotropic, which is unlikely in most realistic scenarios. However, we believe that even in their current form, the proposed methods provide useful tools for studying the fundamental structure of latent spaces, including those of higher dimension as we have hinted at in section \ref{subsec:NLP}, as well as pointing in the directions of promising future research. While lacking some of the desirable quantitative properties of precise Riemannian metrics, our treatment opens up new possibility for qualitative analysis and transformation of latent spaces.
We also believe that they can be effectively used for advanced data visualizations. Further, researchers in fields such as computer graphics and cosmology have worked on similar problems for a long time and developed highly efficient methods for computations in higher dimensions, particularly if this high dimensional space is very sparsely populated as is the case for most high dimensional latent spaces. Applying methods from these fields could drastically improve the applicability of the heuristic metric calculations, as well as the transformation. 

%\subsubsection*{Acknowledgments}

%We acknowledge stuff.
%\bibliographystyle{unsrt}
\bibliography{refs2.bib}

\begin{thebibliography}{10}

\bibitem{Kingma2013-cu}
Diederik~P. Kingma and Max Welling.
\newblock Auto-encoding variational bayes.
\newblock {\em CoRR}, abs/1312.6114, 2013.

\bibitem{Burda2015-qb}
Yuri Burda, Roger~B. Grosse, and Ruslan Salakhutdinov.
\newblock Importance weighted autoencoders.
\newblock {\em CoRR}, abs/1509.00519, 2015.

\bibitem{NIPS2014_5423}
Ian Goodfellow, Jean Pouget-Abadie, Mehdi Mirza, Bing Xu, David Warde-Farley,
  Sherjil Ozair, Aaron Courville, and Yoshua Bengio.
\newblock Generative adversarial nets.
\newblock In Z.~Ghahramani, M.~Welling, C.~Cortes, N.~D. Lawrence, and K.~Q.
  Weinberger, editors, {\em Advances in Neural Information Processing Systems
  27}, pages 2672--2680. Curran Associates, Inc., 2014.

\bibitem{Chen2017-pk}
Nutan Chen, Alexej Klushyn, Richard Kurle, Xueyan Jiang, Justin Bayer, and
  Patrick van~der Smagt.
\newblock Metrics for deep generative models.
\newblock In {\em International Conference on Artificial Intelligence and
  Statistics, {AISTATS} 2018, 9-11 April 2018, Playa Blanca, Lanzarote, Canary
  Islands, Spain}, pages 1540--1550, 2018.

\bibitem{DBLP:journals/corr/abs-1711-08014}
Hang Shao, Abhishek Kumar, and P.~Thomas Fletcher.
\newblock The riemannian geometry of deep generative models.
\newblock {\em CoRR}, abs/1711.08014, 2017.

\bibitem{Arvanitidis2017-oi}
Georgios Arvanitidis, Lars~Kai Hansen, and S{\o}ren Hauberg.
\newblock Latent space oddity: on the curvature of deep generative models.
\newblock {\em CoRR}, abs/1710.11379, October 2017.

\bibitem{Hadjeres2017-ws}
Ga{\"{e}}tan Hadjeres, Frank Nielsen, and Fran{\c{c}}ois Pachet.
\newblock {GLSR-VAE:} geodesic latent space regularization for variational
  autoencoder architectures.
\newblock In {\em 2017 {IEEE} Symposium Series on Computational Intelligence,
  {SSCI} 2017, Honolulu, HI, USA, November 27 - Dec. 1, 2017}, pages 1--7,
  2017.

\bibitem{featuremetric}
Samuli Laine.
\newblock Feature-based metrics for exploring the latent space of generative
  models.
\newblock {\em International Conference on Learning Representations}, 2018.

\bibitem{Peste}
Alexandra Peste, Luigi Malag{\`o}, and Septimia S{\^a}rbu.
\newblock An explanatory analysis of the geometry of latent variables learned
  by variational {Auto-Encoders}.
\newblock {\em NIPS, Bayesian Deep Learning Workshop}, 2017.

\bibitem{DBLP:journals/corr/ChoMGBSB14}
Kyunghyun Cho, Bart van Merrienboer, {\c{C}}aglar G{\"{u}}l{\c{c}}ehre, Fethi
  Bougares, Holger Schwenk, and Yoshua Bengio.
\newblock Learning phrase representations using {RNN} encoder-decoder for
  statistical machine translation.
\newblock {\em CoRR}, abs/1406.1078, 2014.

\bibitem{DBLP:journals/corr/BahdanauCB14}
Dzmitry Bahdanau, Kyunghyun Cho, and Yoshua Bengio.
\newblock Neural machine translation by jointly learning to align and
  translate.
\newblock {\em CoRR}, abs/1409.0473, 2014.

\bibitem{NIPS2014_5346}
Ilya Sutskever, Oriol Vinyals, and Quoc~V Le.
\newblock Sequence to sequence learning with neural networks.
\newblock In Z.~Ghahramani, M.~Welling, C.~Cortes, N.~D. Lawrence, and K.~Q.
  Weinberger, editors, {\em Advances in Neural Information Processing Systems
  27}, pages 3104--3112. Curran Associates, Inc., 2014.

\bibitem{476cadd89dec4e0ab01d9dc59e1222c7}
Samuel~R. Bowman, Luke Vilnis, Oriol Vinyals, Andrew~M. Dai, Rafal
  J{\'{o}}zefowicz, and Samy Bengio.
\newblock Generating sentences from a continuous space.
\newblock In {\em Proceedings of the 20th {SIGNLL} Conference on Computational
  Natural Language Learning, CoNLL 2016, Berlin, Germany, August 11-12, 2016},
  pages 10--21, 2016.

\bibitem{Yang2017-dj}
Zichao Yang, Zhiting Hu, Ruslan Salakhutdinov, and Taylor Berg{-}Kirkpatrick.
\newblock Improved variational autoencoders for text modeling using dilated
  convolutions.
\newblock In {\em Proceedings of the 34th International Conference on Machine
  Learning, {ICML} 2017, Sydney, NSW, Australia, 6-11 August 2017}, pages
  3881--3890, 2017.

\bibitem{DBLP:journals/corr/OordDZSVGKSK16}
A{\"{a}}ron van~den Oord, Sander Dieleman, Heiga Zen, Karen Simonyan, Oriol
  Vinyals, Alex Graves, Nal Kalchbrenner, Andrew~W. Senior, and Koray
  Kavukcuoglu.
\newblock Wavenet: {A} generative model for raw audio.
\newblock {\em CoRR}, abs/1609.03499, 2016.

\bibitem{jurafsky2000speech}
Dan Jurafsky.
\newblock {\em Speech \& language processing}.
\newblock Pearson Education International, 2000.

\bibitem{MacKay:2002:ITI:971143}
David J.~C. MacKay.
\newblock {\em Information Theory, Inference \& Learning Algorithms}.
\newblock Cambridge University Press, New York, NY, USA, 2002.

\bibitem{b3ad6758e4074325b34caeadce5a91b5}
DFL Dorling.
\newblock {\em Area cartograms: their use and creation}.
\newblock Environmental Publications, University of East Anglia, 1996.

\bibitem{Gastner2004-iv}
Michael~T Gastner and M~E~J Newman.
\newblock Diffusion-based method for producing density-equalizing maps.
\newblock {\em Proc. Natl. Acad. Sci. U. S. A.}, 101(20):7499--7504, May 2004.

\bibitem{Gastner2018-fr}
Michael~T. Gastner, Vivien Seguy, and Pratyush More.
\newblock Fast flow-based algorithm for creating density-equalizing map
  projections.
\newblock {\em CoRR}, abs/1802.07625, 2018.

\bibitem{scikit-learn}
F.~Pedregosa, G.~Varoquaux, A.~Gramfort, V.~Michel, B.~Thirion, O.~Grisel,
  M.~Blondel, P.~Prettenhofer, R.~Weiss, V.~Dubourg, J.~Vanderplas, A.~Passos,
  D.~Cournapeau, M.~Brucher, M.~Perrot, and E.~Duchesnay.
\newblock Scikit-learn: Machine learning in {P}ython.
\newblock {\em Journal of Machine Learning Research}, 12:2825--2830, 2011.

\end{thebibliography}

\end{document}